%% file: main.tex
\documentclass[10pt,twocolumn,letterpaper]{article}

\usepackage{cvpr}     

\usepackage{graphicx}
\usepackage{amsmath}
\usepackage{amssymb}
\usepackage{booktabs}

\usepackage{times}
\usepackage{epsfig}
\usepackage{float}

\usepackage{enumitem} %
\usepackage{caption}
\usepackage{multirow} %
\usepackage{pifont}
\usepackage{placeins}

\usepackage[dvipsnames]{xcolor}
\definecolor{my_pink}{HTML}{FF33D7}
\definecolor{my_green}{HTML}{0AF718}
\definecolor{cvprblue}{rgb}{0.21,0.49,0.74}

\usepackage[pagebackref,breaklinks,colorlinks,citecolor=cvprblue]{hyperref}

\definecolor{somegray}{rgb}{0.5, 0.5, 0.5}
\newcommand{\darkgrayed}[1]{\textcolor{somegray}{#1}}
\makeatletter
\newcommand*\titleheader[1]{\gdef\@titleheader{#1}}
\AtBeginDocument{%
  \let\st@red@title\@title
  \def\@title{%
    \vskip-4em
    \bgroup\normalfont\large\centering\@titleheader\par\egroup
    \vskip1.5em\st@red@title}
}
\makeatother

\titleheader{\darkgrayed{This paper has been accepted for publication at the\\
IEEE Conference on Computer Vision and Pattern Recognition Workshops (CVPRW), Seattle, 2024.}}

\title{A Hybrid ANN-SNN Architecture \\for Low-Power and Low-Latency Visual Perception\\ }

\author{Asude Aydin \hspace*{20pt} Mathias Gehrig \hspace*{20pt} Daniel Gehrig \hspace*{20pt} Davide Scaramuzza
\vspace{0.2cm} \\
Robotics and Perception Group, University of Zurich, Switzerland}

\begin{document}

\maketitle

\input{content/00_abstract}
\vspace{-\baselineskip}

\noindent
\textbf{Code:} \url{https://github.com/uzh-rpg/hybrid_ann_snn}
\input{content/01_intro.tex}

\vspace{-5pt}
\input{content/02_related_modified.tex}
\vspace{-5pt}
\input{content/03_methods.tex}

\vspace{-5pt}
\input{content/04_experiments.tex}

\vspace{-5pt}
\input{content/05_conclusion.tex}

\vspace{-5pt}
\section{Acknowledgement}
This work was supported by the National Centre of Competence in Research (NCCR) Robotics (grant agreement No. 51NF40-185543) through the Swiss National Science Foundation (SNSF), and the European Research Council (ERC) under grant agreement No. 864042 (AGILEFLIGHT).

\newpage
\input{content/appendix}

{\small
\bibliographystyle{ieee_fullname}
\bibliography{references}
}

\end{document}

%% file: content/00_abstract.tex
\begin{abstract}
Spiking Neural Networks (SNNs) are a class of bio-inspired neural networks that promise to bring low-power and low-latency inference to edge-devices through the use of asynchronous and sparse processing. 
However, being temporal models, SNNs depend heavily on expressive states to generate predictions on par with classical artificial neural networks (ANNs).
These states converge only after long transient time periods, and quickly decay in the absence of input data, leading to higher latency, power consumption, and lower accuracy.  
In this work, we address this issue by initializing the state with an auxiliary ANN running at a low rate. The SNN then uses the state to generate predictions with high temporal resolution until the next initialization phase. Our hybrid ANN-SNN model thus combines the best of both worlds: It does not suffer from long state transients and state decay thanks to the ANN, and can generate predictions with high temporal resolution, low latency, and low power thanks to the SNN.   
We show for the task of event-based 2D and 3D human pose estimation that our method consumes 88\% less power with only a 4\% decrease in performance compared to its fully ANN counterparts when run at the same inference rate. Moreover, when compared to SNNs, our method achieves a 74\% lower error.
This research thus provides a new understanding of how ANNs and SNNs can be used to maximize their respective benefits.
\end{abstract}

%% file: content/01_intro.tex
\section{Introduction}\label{sec:intro}

\begin{figure}
    \centering
    \includegraphics[width=1\linewidth]{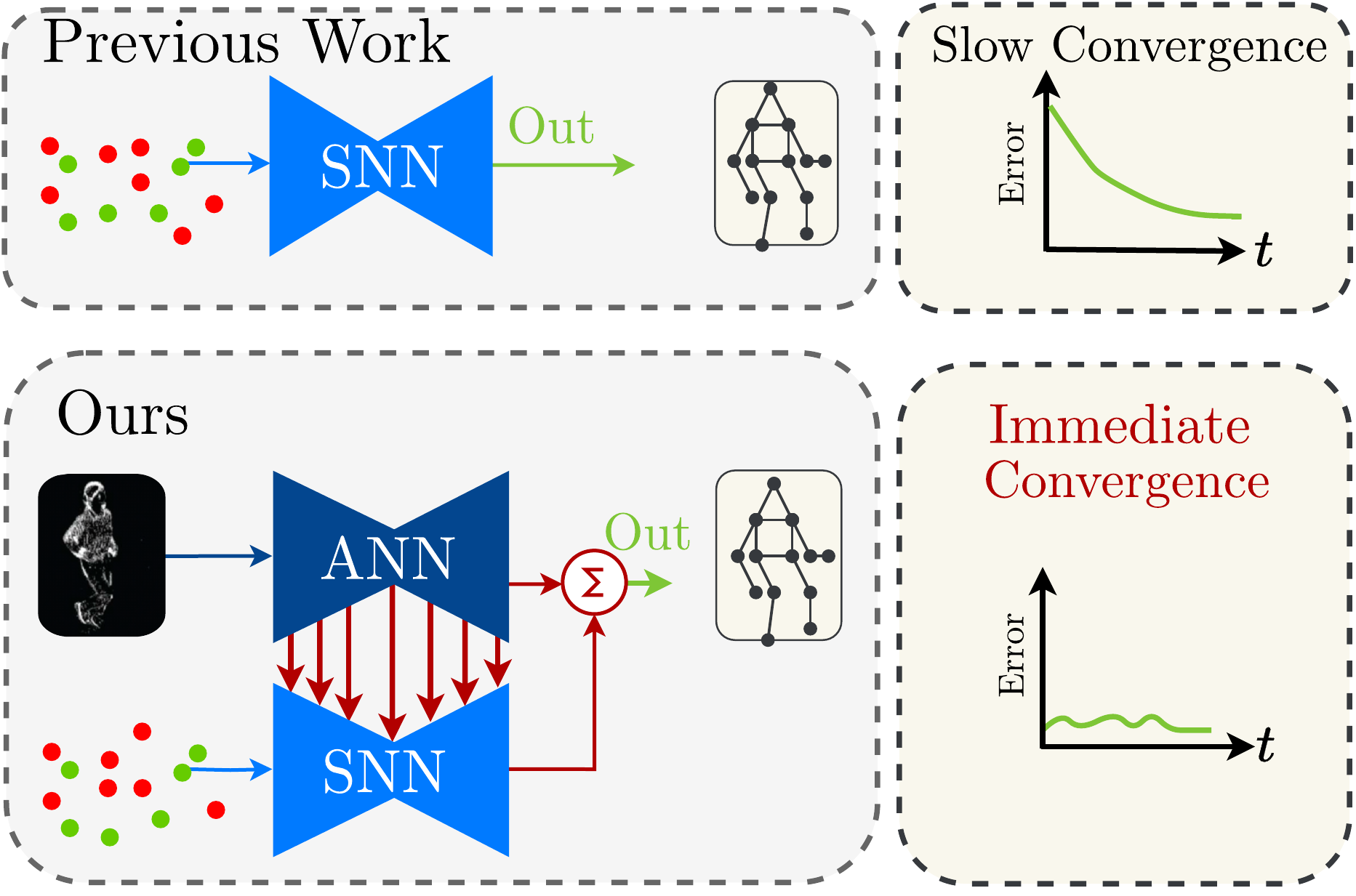}
    \caption{Spiking Neural Networks (SNNs, top) are prone to long transient periods and state decay in the absence of input data, leading to lower accuracy and higher latency and power consumption. In this work (bottom), we solve this with an auxiliary artificial neural network (ANN) that initializes the SNN states at low rates. Our resulting hybrid architecture is simultaneously accurate and maintains the low-power and low-latency aspect of SNNs.  \vspace{-\baselineskip} }
    \label{fig:intro_fig}
\end{figure}

Recent breakthroughs in deep learning have led to accelerating progress on a wide range of computer vision tasks. As this progress speed-ups, practitioners are moving to deeper and deeper models in the pursuit of higher task performance. 
However, this trend comes at a cost: Today's large-scale models require increasing amounts of power, which limits their adoption in power-constrained scenarios. 

Low-power computation is a crucial requirement for applications running on edge devices and can make the difference between changing the battery of IoT devices once a day or once a year or greatly increasing the mission time of autonomous robots with power-constrained hardware. 

Spiking Neural Networks (SNNs) are a novel brain-inspired way to process visual signals, which are orders of magnitude more efficient than their Artificial Neural Network (ANN) counterparts. Instead of processing inputs as synchronous, analog-valued tensor maps, they are dynamical systems that process data as sparse spike trains. Moreover, when deployed on neuromorphic processors, SNNs function asynchronously in an activity-driven fashion, enabling fast inference and low power consumption. 

Although previous work has demonstrated SNN applications on a wide range of tasks, they are still limited in their performance due to two shortcomings: First, they are difficult to train due to the non-differentiability of spikes~\cite{surrogate_gradient} and the vanishing-gradient problem~\cite{spiking_vanishing, pascanu2013difficulty}. Second, they require long time windows to converge and match the accuracy of ANNs~\cite{li2021free, fang2021deep} and are prone to decay in the absence of input data.
This is because SNNs, being temporal models, depend on an expressive state to generate predictions on par with classical ANNs. 
However, this state takes time to converge: Membrane potentials (i.e., states) in the SNN layers need to charge over time, then cross the firing threshold, and finally emit spikes that charge the next layer. In deep SNNs, this charging time generates a long delay, which increases latency and energy consumption due to the need for more iterations.

In this work, we solve this issue by using an auxiliary ANN to initialize the SNN states periodically at low rates, thus eliminating the need for the SNN states to converge. 
We then use the SNN to generate predictions and propagate the state forward until the next initialization step. The resulting predictions (i) have a high rate, (ii) experience a boost in accuracy due to a well-initialized state, and (iii) maintain the low-power property of SNNs. Crucially, the auxiliary ANN only requires little power due to the low rate of state initialization. The resulting hybrid model thus combines the advantages of both ANNs and SNNs. 

We evaluate our method on the tasks of 2D and 3D human pose estimation (HPE) using events from an event camera, where we show consistently that our method reduces the power consumption of standard ANNs by 88\% while only achieving a 4\% error increase. Instead, when compared to SNNs, our method achieves 74\% lower error. 

In summary:
\begin{enumerate}
    \item We propose an energy-efficient, low-latency hybrid ANN-SNN architecture, where the ANN is tasked with initializing the SNN states at low frequency, thus overcoming the limitations of both ANNs and SNNs. 
    \item We show for the task of event-based HPE that this method achieves a balance between being accurate and power efficient. Compared to standard ANNs, it achieves significant improvements in terms of energy consumption and update rate while only experiencing a slight decrease in accuracy. 
    \item The proposed method naturally supports frame-based input, such as RGB images, which further improves the accuracy.
\end{enumerate}

%% file: content/02_related_modified.tex
\section{Related work}\label{sec:related} 
\subsection{Hybrid ANN-SNN Architectures}\label{subsec:related_hybrid}

In recent years, there has been a growing interest in exploring the potential benefits of combining Artificial Neural Networks (ANNs) and Spiking Neural Networks (SNNs)~\cite{hybrid_dashnet,spikeflownet,hybrid_obj_detection,hybrid_top_down,hybrid_units}. 
Different combination strategies have been explored for a variety of tasks.

A group of work employs the strategy of processing the accumulated spike train of SNNs with ANNs \cite{hybrid_obj_detection, hybrid_top_down, spikeflownet}.
In these works, the SNN is used as an efficient encoder of spatio-temporal event data from an event camera.
The output of the SNN is accumulated to summarize the temporal dimension before the ANN processes the accumulated features \cite{hybrid_obj_detection, spikeflownet}.
Liu et al. \cite{hybrid_top_down} extend this idea for object classification and use a feedback loop from the ANN to the input of the SNN.
The downside of the aforementioned approaches is that they need to execute a full forward pass of the ANN to extract results, which results in high power consumption and high computational latency. In contrast, our SNN directly updates the output of the ANN such that we do not need to execute the ANN for every iteration, thus retaining the low-power property.

A second line of work uses a strategy where the output of the independently operating SNN and ANN is fused \cite{lele2022bio, hybrid_dashnet, hybrid_units}.
This approach is especially suitable for multimodal processing of frame-based video and event data from event cameras.
The ANN is tasked with extracting features from the frames while the SNN processes events directly.
Finally, the output of both networks are fused based on heuristics \cite{lele2022bio}, temporal filtering \cite{hybrid_dashnet}, or accumulation based on the output of the ANN \cite{hybrid_units}.
However, these methods do not address the convergence of SNN's and also do not share features between networks, making their fusion shallow. By contrast, our approach not only reuses the output of the ANN \cite{hybrid_units} but also reuses the features from the ANN to initialize the SNN states. The initialization of SNN states drastically improves the performance and convergence of the SNN.

\subsection{Human Pose Estimation}\label{subsec:related_hpe}

\paragraph{Frame-based} Human pose estimation (HPE) is the task of estimating the 2D or 3D locations of body joints from a single image or video.
Current techniques for 3D HPE involve reconstructing the 3D pose from either single~\cite{chen2017_3d_singleframe, mehta2018single, pavlakos2017coarse, tome2017lifting, zhou2017towards, monocular_hpe_survey} or multiple~\cite{kocabas2019self, hofmann2012multi,elhayek2015efficient,rhodin2016general} camera views.
To estimate the 3D pose from multiple views, the traditional approach involves predicting the 2D pose in each view and using the camera characteristics and positions to triangulate it into the world coordinate~\cite{hpe_triangulation}. 
Alternatively, newer approaches include triangulation with neural networks~\cite{iskakov2019learnable, liu2020attention, liu2021enhanced, katircioglu2018learning} or direct regression of the 3D pose~\cite{yao2019densebody, rhodin2016general}.
Both single-view and multi-view approaches can be improved by using multiple frames to extract temporal information that can help disambiguate joint locations over time and reduce jitters~\cite{mehta2017vnect, hofmann2012multi, elhayek2015efficient,rhodin2016general,katircioglu2018learning}.
\vspace*{-0.3cm}
\paragraph{Event-based} Recently, human pose estimation with event cameras has gained traction due to their inherent ability to filter out temporally redundant information like the background~\cite {eventcap, eventhpe, dhp19, tore_volume, lifting_monocular_events}. These works adopt one of two main approaches.
The first direction of work utilizes volumetric human body models \cite{smpl} to estimate both the 3D pose and shape of the human body \cite{eventcap, eventhpe}.
EventCap~\cite{eventcap} and EventHPE~\cite{eventhpe} use a low-dimensional human shape representation called SMPL~\cite{smpl} to enable end-to-end shape and pose estimation from images and events.
The second approach, in contrast, focuses on extracting the pose information using a skeleton body model \cite{lifting_monocular_events, dhp19, tore_volume}. 
Scarpellini et al.~\cite{lifting_monocular_events} use events from a single camera view and predict the 3D pose with a Convolutional Neural Network (CNN).
Calabrese et al.~\cite{dhp19} and Baldwin et al.~\cite{tore_volume} estimate 2D joint locations using a CNN architecture and perform triangulation to obtain the 3D pose.
Different from previous work, we target low-power inference for human pose estimation using a hybrid ANN-SNN architecture.
Our approach benefits from the high accuracy of ANNs and low power consumption of SNNs to improve the accuracy to power tradeoff.

%% file: content/03_methods.tex
\section{Methodology}\label{sec:methods}
\begin{figure}
    \centering
    \includegraphics[width=\linewidth]{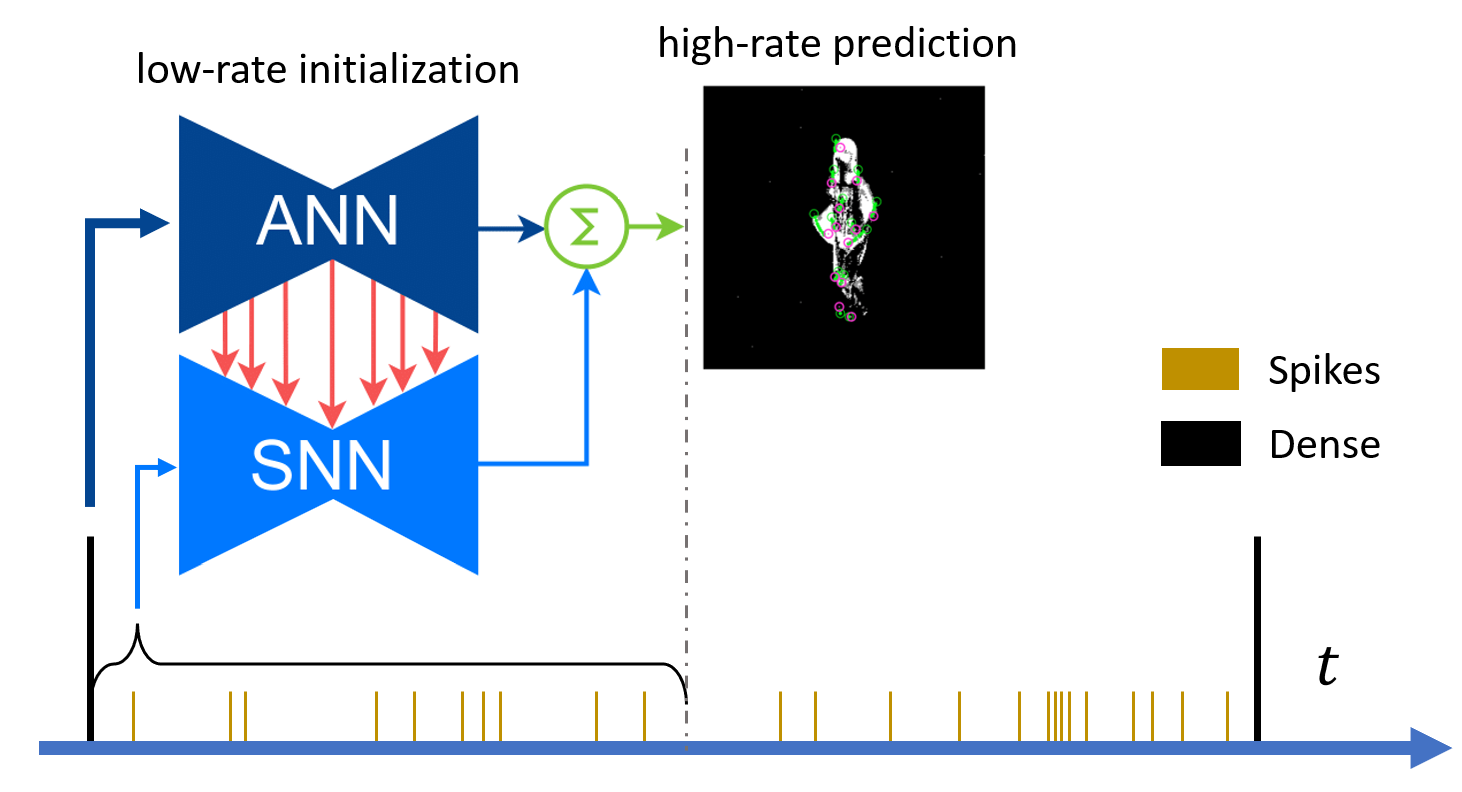}
    \caption{\textbf{Overview of our method}. Our method processes inputs as dense and spike-based representations. The ANN uses the dense representation to perform state and output initialization at low rates. The SNN then uses spikes to generate high-rate outputs until the next dense input.  \vspace{-\baselineskip} }
    \label{fig:method_overview}
\end{figure}

\begin{figure*}[hbt]
\centering
  \includegraphics[width=1.0\textwidth]{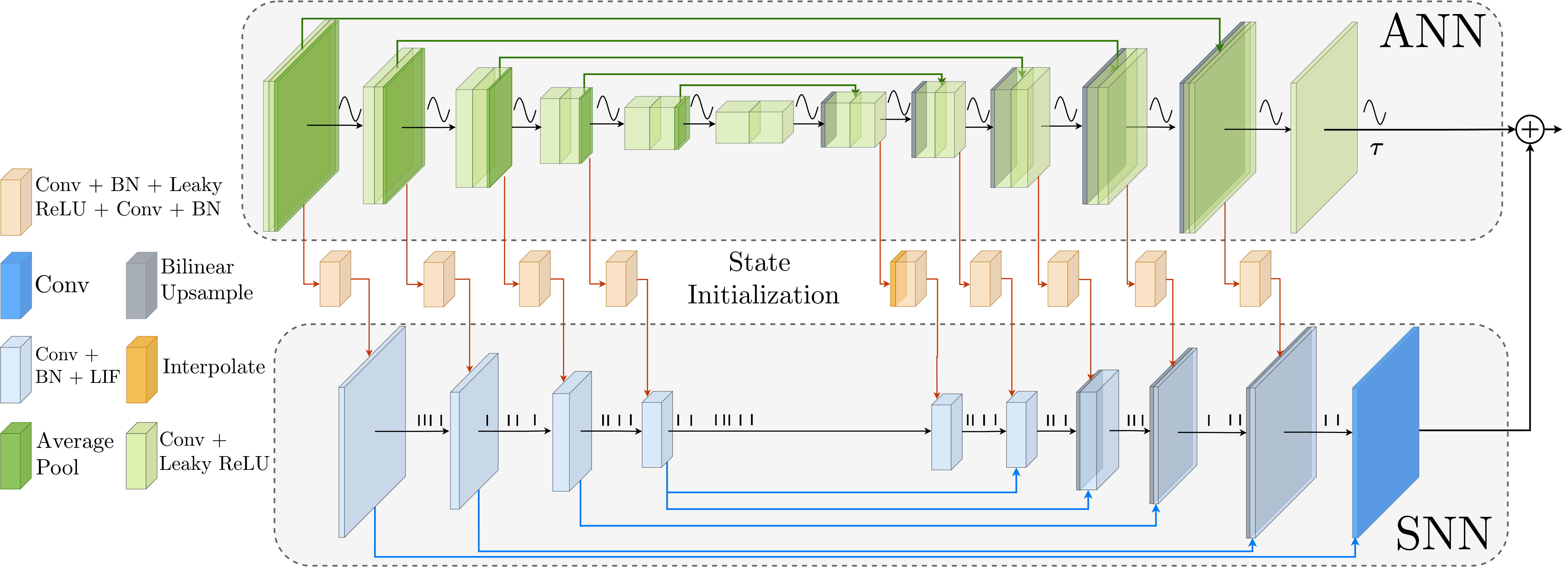}
  \caption{\textbf{Hybrid ANN - SNN architecture.} The ANN (upper row of blocks) is fed with past events at time step $t_0$, where an initial output is predicted, and states of spiking neurons are initialized (orange blocks). Events of duration $\Delta T$ are fed sequentially to the SNN (lower row of blocks) for high-rate updates of the prediction.}
  \label{fig:hybrid_arch}
\end{figure*}

An overview of our approach is depicted in Fig.~\ref{fig:method_overview}. 
Our approach employs a slow-fast~\cite{feichtenhofer2019slowfast, censi2014low} two-branch design where feature extraction alternates between an expensive but infrequent ANN stage and a frequent cost-effective SNN stage.
In our hybrid ANN-SNN approach, the ANN is utilized to accurately predict joint locations based on prior events and to simultaneously initialize the spiking neuron states. 
This sets the stage for high-frequency, low-latency updates using the SNN, where events of duration $\Delta T$ are sequentially fed to the SNN.
After the sequence duration $T$, the process is repeated where a new prediction is made by the ANN, and states of spiking neurons are re-initialized.
The integration of the ANN at low rates and the SNN at high rates enables precise predictions to be made with low-latency while maintaining energy efficiency.
The following provides a detailed explanation of our hybrid model and its constituent steps.

\subsection{Preliminaries}
Our method takes as input a sequence of spike-based and dense representations. The dense representation can be an image, if synchronized and aligned with events, or any dense event representation computed from raw events. For the remainder of this section, we let $Y_i$ be the dense representation at time $t_i$. In this work, we opted for stacked 2D histograms~\cite{Gehrig19iccv} in case of event data. They are computed by stacking $N=10$ two-channel histograms~\cite{Maqueda18cvpr} from a total of 7,500 events.
We then consider raw binary events up to time $t$ denoted as $X(t)$ with $t > t_i$:
\begin{equation}
\label{eq:spike_train}
    X(t) = \sum_{j | t_j < t} p_j\delta(t-t_j).
\end{equation}
In general, $t_j$ is the timestamp of the $j$-th events, and $p_j\in\mathbb{R}^2$ is the event polarity converted to a one-hot vector.
The ANN, $F_\text{ANN}$, processes the dense event representation and predicts both the output $o$ at time $t$ as well as the initial SNN states $\{s^k\}_{k=1}^L$ for all $L$ layers.
The SNN, $F_\text{SNN}$, then processes the incoming event stream $X$ to continuously update the prediction $o$.
The following equations summarize this process:
\begin{align}
    \{s^k_i\}_{k=1}^L, o_i &= F_\text{ANN}(Y_i)\\
    o(t) &= F_\text{SNN}(t; X, \{s^k_i\}_{k=1}^L, o_i).
\end{align}
Here, $\{s^k_i\}_{k=1}^L$ denote the membrane potential of the SNNs at layers $k=1,...,L$ for timestamp $t_i$. The variable $o_i$ denotes the output map for timestamp $t_i$. Finally, $o(t)$ denotes the human pose estimates at time $t$. It is generated by using the initialization $o_i$ and integrating the output of the SNN onto it.  
In summary, the task of the SNN is to incrementally update the initial prediction that the ANN provides.
While the ANN is a standard U-Net~\cite{Ronneberger15icmicci}, the SNN can be interpreted as a continuous-time model that takes a function $X$ (see Eq.~\eqref{eq:spike_train}) as input and generates a prediction at any time $t$. Next, we will go into more detail on how this model works. 

\subsection{Spiking Neural Networks}
SNNs model individual neurons at layer $k$ as dynamical systems that update their membrane potential $V^k$ by integrating a series of input spikes in a learnable way. When their membrane potential exceeds a threshold, it generates spikes which are then transmitted to the next layer, followed by some resetting of the membrane potential. In our work, we use the Leaky Integrate \& Fire (LIF) neuron model~\cite{gerstner_neuronal_dynamics}.
The sub-threshold dynamics of a LIF neuron are defined as
\begin{equation}
    \label{eqn:lif}
    \tau \frac{dV^k(t)}{dt} = -(V^k(t) - V_{rest}) + X^k(t).
\end{equation}
Here $V^k(t)$ represents the neuron's membrane potential at time t, and layer $k$, $V_{rest}$ is the resting potential of the neuron, $X^k(t)$ denotes the integrated spike train at time t, and $\tau$ is the membrane time constant. After the membrane potential reaches the firing threshold $V_{th}$, a spike is emitted, and the membrane potential is immediately reset back to its resting potential.   

While conventionally, membrane potentials are initialized at 0 for all neurons and layers; we use the ANN to initialize these potentials in this work. We thus modify Eq.~\ref{eqn:lif} by adding a boundary condition at time $t_i$ for each layer:
\begin{align}
    \label{eqn:lif_modified}
    \tau \frac{dV^k(t)}{dt} &= -(V^k(t) - V_{rest}) + X^k(t). \\
    \nonumber\text{subj. to: } V^k(t_i) &= s^k_i
\end{align}
where $s^k_i$ are the activation initialization maps generated by the ANN. As will be shown later, this small change has a major impact on the SNN behavior since it mitigates delays due to convergence in Eq.~\ref{eqn:lif} and improves performance overall by providing a well-initialized state. Next, we will discuss how we emulate such a continuous dynamical system on conventional hardware. 

\subsection{Discretization and Training}
To train our ANN-SNN model, we need to convert the SNN into a recurrent network. This is typically done by applying a forward Euler approximation to the differential Eq.~\eqref{eqn:lif_modified}. Eq.~\eqref{eqn:lif_discrete} shows the discretized sub-threshold dynamics and the spiking mechanism where $H(.)$ is the Heaviside step function with $S_{t}$ denoting the spike output.
\begin{align}
\label{eqn:lif_discrete}
\begin{split}
    & V^k_{t} = V^k_{t-1} + \frac{1}{\tau}(X^k_{t} - (V^k_{t-1} - V_{rest})) \\
    & S^k_{t} = H(V^k_{t} - V_{th})  \\
    \text{subj. to: } & V^k_0 =s^k_i
\end{split}
\end{align}
In the discretized version, time $t$ takes integer values and starts at index 0, which previously corresponded to the timestamp $t_i$. 
To emulate the resetting behavior, we apply soft resets, also known as Residual Membrane Potential Neurons~\cite{rmp_neurons}, which reduce the potential by the amount of the threshold value. This allows the residual potential to be re-used at the next steps, resulting in reduced information loss.
Using soft reset neurons has the effect that potential values can be initialized outside the range $[V_{rest}, V_{thr}]$. This allows extreme cases such as dead neurons or always ON neurons.

Eqs.~\eqref{eqn:lif_discrete} can be interpreted as a recurrent neural network that can be unrolled over multiple forward Euler steps and then trained using backpropagation through time~\cite{bptt}. However, a challenge in training the above spiking neural networks lies in its use of the Heaviside step function $H$, which is not differentiable. However, this problem can be addressed by using surrogate gradients~\cite{surrogate_gradient}, i.e., replacing the gradient of the Heaviside function with the approximate
\begin{equation}
    H'(x) \approx \frac{1}{1+(\pi x)^2} 
\end{equation}
For more details on SNN training, see \cite{surrogate_gradient}.

\subsection{Network Details}
The hybrid network details are illustrated in Fig.~\ref{fig:hybrid_arch}. 
For the ANN (top row), we use a U-Net structure~\cite{unet}, adapted from Super SloMo~\cite{super_slomo}. It has a total of 23 layers, comprising a prediction layer, five encoders, and five decoders concatenated with skip connections at the same spatial resolution. 
The SNN (bottom row) is a variant of the U-Net architecture~\cite{unet}, modified from EVSNN~\cite{evsnn}. 
The architecture consists of 10 layers made up of a prediction layer, residual block, four encoder, and four decoder layers. 
At every timestep, events are presented to the network in two channels for each polarity. 
The network is trained with a discretization step of 10 ms. 
At the output, we use a simple convolutional layer and integrator, which allows predicting analog heatmaps.
Each SNN layer is initialized from ANN state initialization modules depicted with orange blocks in Fig.~\ref{fig:hybrid_arch}.
The initialization module reuses the ANN U-Net features and predicts the membrane potential of the SNN spiking neurons.
The initialization modules consist of only two convolutional layers followed by batch normalization.
An ablation study of the state initialization architectures can be found in Sec.~\ref{subsec:results_stateinit_ablation}.
Additional network details about the ANN and SNN are given in the appendix.

%% file: content/04_experiments.tex
\section{Experiments}\label{sec:experiments}
We evaluate our model on two publicly available event-based human pose estimation datasets, DHP19~\cite{dhp19} and Event-Human3.6M~\cite{lifting_monocular_events}. 
Sec.~\ref{sec:exp:setup} starts off with details about the datasets, metrics, and training. We then demonstrate the effectiveness of our hybrid model in reducing power consumption while boosting accuracy in the ablation study section (Sec.~\ref{sec:exp:ablations}). We continue with a spike activity analysis in Sec.~\ref{sec:spike_activity} and comparison with state-of-the-art event-based human pose estimators in Sec.~\ref{sec:exp:sota} before concluding with a limitation analysis of our approach in Sec.~\ref{sec:exp:limitation}. Further analysis and insights on state initializations can be found in the Appendix.\\

\vspace{-0.3cm}
\subsection{Setup}
\label{sec:exp:setup}
\paragraph{Datasets}
The DHP19 dataset \cite{dhp19} is a real-world event camera dataset recorded with 4 synchronized DAVIS346 cameras. Overall, the dataset features 17 different subjects performing a total of 33 movement patterns. DHP19 consists of 556 sequences with labels from a motion capture system at 100 Hz. 
With this dataset, we analyze our pure event-based implementation using real event data.

The Event-Human3.6M \cite{lifting_monocular_events} originates from the Human3.6M dataset~\cite{human36m} and uses an event simulator \cite{Gehrig_2020_CVPR} to generate synthetic event data.
The dataset features 11 subjects and 17 different activities that are more complex than the movements in the DHP19 dataset.
Human3.6M images are captured with 4 synchronized high-resolution cameras and labels with a motion capture system at 50 Hz. 
Our experiments on Event-Human3.6M examine whether our approach also performs well on more complex motion patterns at the cost of using synthetic event data.

\paragraph{Metrics} We measure the \emph{accuracy} using the Mean Per Joint Position Error (MPJPE). For each joint, the Euclidean distance between the predicted and ground truth positions of a joint is calculated.
The MPJPE score is computed as the mean of these errors across all joints in the skeleton body model.
This score is defined for both 2D skeleton estimations (in pixels) and 3D estimation (in millimeters) as: 

\begin{equation}
    \centering
    \label{eqn:mpjpe}
    \text{MPJPE} = \frac{1}{J} \sum_{i = 1}^J \|{x_i - \hat{x}_{i}}\|,
\end{equation}

where $J$ is the number of joints, $x_{i}$ is the ground truth, and $\hat{x}_{i}$ is the estimation of the joint in 2D or 3D space.
\vspace{-5pt}
\paragraph{Energy Consumption} depends on computation and data movement, which includes memory access.
However, due to hardware dependencies, quantifying memory access and data movement is challenging.
Therefore, we measure energy consumption by computing the total number of multiply-accumulate (MAC) and accumulate (AC) operations used by a method. While ANNs perform dense MAC operations, SNNs perform sparse AC operations as a result of the binary nature of spikes. In most technologies, the addition operation is less costly than the multiplication operation. For 7nm CMOS technology, one 32-bit MAC operation uses 1.69 pJ, while one AC only uses 0.38 pJ~\cite{ieee_operation_cost_7nm}, which are the values we use to calculate the power consumption of all methods. For ANNs, we compute the total number of MACs throughout the layers as $k^2 W_o H_o C_i C_o$, where we use the kernel size $k$, output dimension $W_o\times H_o$ and input and output channel dimension $C_i$ and $C_o$. For SNNs, we count the total number of ACs as the above value multiplied by the average spiking activity $\zeta^{l}\in[0,1]$. It is the ratio of the total number of spikes in layer $l$ over all timesteps to the total number of neurons in layer $l$~\cite{spikeact1, spikeact2, spikethrift}. In these calculations, we assume every spike consumes constant energy~\cite{spike_consumption}.

\input{content/ablation_figure.tex}
\vspace{-5pt}
\paragraph{Training Details}\label{subsec:methods_training}
For SNN training, we use the Spiking Jelly framework~\cite{spikingJelly}, an open-source deep learning framework based on PyTorch~\cite{pytorch}.
The network is unrolled for all time steps to perform BPTT~\cite{bptt} on the average loss of the sequence, $L_{\text{avg}}$. 
We adopt the loss function from~\cite{Calabrese19cvprw}, which computes the difference between heatmaps generated by our method and 3D joint labels projected to pixel space. 
For each 2D joint, ground-truth heatmaps are generated by setting the joint location pixel to 1 in a zero-filled 2D tensor at the input's spatial resolution. 
Gaussian blurring with a filter size of 11 and a standard deviation of 2 pixels is applied to each heatmap to aid learning. 
The loss is computed between predicted and ground truth heatmaps using the Mean Square Error (MSE) and averaged over several timesteps
\begin{equation}
    \centering
    \label{eqn:mse}
    L_\text{avg} = \frac{1}{JT}\sum_{t=1}^T \sum_{i = 1}^J ({o_{ti} - \hat{o}_{ti}})^{2},
\end{equation}
All experiments were trained on a single GPU, Quadro RTX 8000, for approximately 72 hours.
We first train an ANN with the learning rate 1e-4 of batch size 8 for 60,000 iterations. 
We then train the hybrid model by freezing the ANN weights and only training the SNN. 
Hybrid experiments were trained for 280,000 iterations with an Adam optimizer~\cite{adam}, batch size 2, and learning rate 5e-5 with the neuronal time constant, firing threshold, and output decay $\tau$ set to 3, 1, and 0.8, respectively. 
Pure SNN experiments were trained for 160,000 iterations with the same parameters, with the exception of the time constant $\tau=2$.

\begin{table}[t!]
\centering
\caption{\textbf{State initialization module ablation study.} Results are reported on the validation set of DHP19 for camera view \#2. The last bin provides the score of the $10^{th}$ time step, and the second column reports the average score of the sequence.}
\label{tab:state_init_ablations}
\resizebox{\columnwidth}{!}{
\begin{tabular}{lcc}
\toprule
\multicolumn{1}{c}{\multirow{2}{*}{\textbf{State Initialization Mappings}}} & \multicolumn{2}{c}{\textbf{ MPJPE (2D) $\downarrow$}} \\ \cline{2-3} 
\multicolumn{1}{c}{}                          & Last Bin & Average \\ \midrule
\multicolumn{1}{c}{-}                         & 9.12     & 7.08    \\ \midrule
Conv + BN + LeakyReLU + Conv + BN + Sigmoid   & 16.59    & 9.83   \\
Conv + BN + LeakyReLU + Conv + BN + LeakyReLU & 6.32    & 6.14   \\
Conv + BN + LeakyReLU + Conv + BN             & \textbf{6.11}    & \textbf{6.02}   \\
Conv + BN + LeakyReLU + Conv                  & 6.22    & 6.23   \\
Conv + BN + LeakyReLU                         & 6.12    & 6.17   \\ \bottomrule
\end{tabular}
}
\end{table}

\subsection{Ablation Studies}
\label{sec:exp:ablations}

This section first examines the two main contributors to the performance of our proposed model, state initialization and output initialization. 
Second, we examine different state initialization architectures.
In these experiments, we run the ANN at 10~Hz and the SNN at 100~Hz with 10~ms of events presented at the input.

\vspace{-10pt}
\subsubsection{Proposed Method}

Fig.~\ref{fig:method_ablations}~reports ablation studies comparing four variants of our method (A-D) in terms of 2D MPJPE on camera view~2 of DHP19. In (E-F), we show the MPJPE over 100 ms of events. 
Model A is a pure SNN with a zero state initialization at time 0. Model B initializes the SNN states from an auxiliary ANN. This ANN processes a dense event representation constructed at time $t=0$ to generate these states, and the SNN then continues predicting within the time interval. Model C does not use state initialization but instead uses the SNN to learn a delta on the ANN prediction at timestamp $t=0$, which we call output initialization. Finally, model D combines the idea of output initialization and state initialization, which is our proposed method.

We see in Fig.~\ref{fig:method_ablations}~(F) that the SNN with states initialized at 0 (A), results in high error due to slow convergence over the time interval shown. One way to solve this is to use output initialization (C), which achieves a low error at the beginning (E) but diverges back to pure-SNN error levels the further away we get from the first ANN prediction. Our proposed way to initialize SNN states via the ANN (B) accelerates the convergence of the SNN and thus leads to stable and low error rates in (E). Finally, adding back the output initialization (D) further reduces the error rate throughout the interval. Crucially, using output initialization upper bounds the method's error rate at $t=0$ to that of the ANN.

\vspace{-0.3cm}
\subsubsection{State Initialization Architecture}\label{subsec:results_stateinit_ablation}

We investigate different blocks for mapping ANN features to initialize membrane potentials (Fig.~\ref{fig:hybrid_arch}, orange blocks) and report the last bin and average bin scores in Tab.~\ref{tab:state_init_ablations}. For the last bin, we report the error achieved after 100 ms of events, and for the average bin score, we average over 10 time steps.

We compare these scores with membrane potentials initialized with zeros reported in the first row. 
In general, we found that putting a batch normalization layer at the end led to the best results. Interestingly, since batch normalization can generate values outside of the range [0,1], it can permanently kill or activate neurons which proved to be beneficial. This can be seen when comparing rows 2 to 4, where adding a range limiting function LeakyReLU or Sigmoid degrades performance. Following Tab.~\ref{tab:state_init_ablations}, we chose Conv + BN + LeakyReLU + Conv + BN.
\subsection{Spike Activity Analysis}
\label{sec:spike_activity}

\begin{figure}[t!]
    \centering
    \includegraphics[width=0.93\columnwidth]{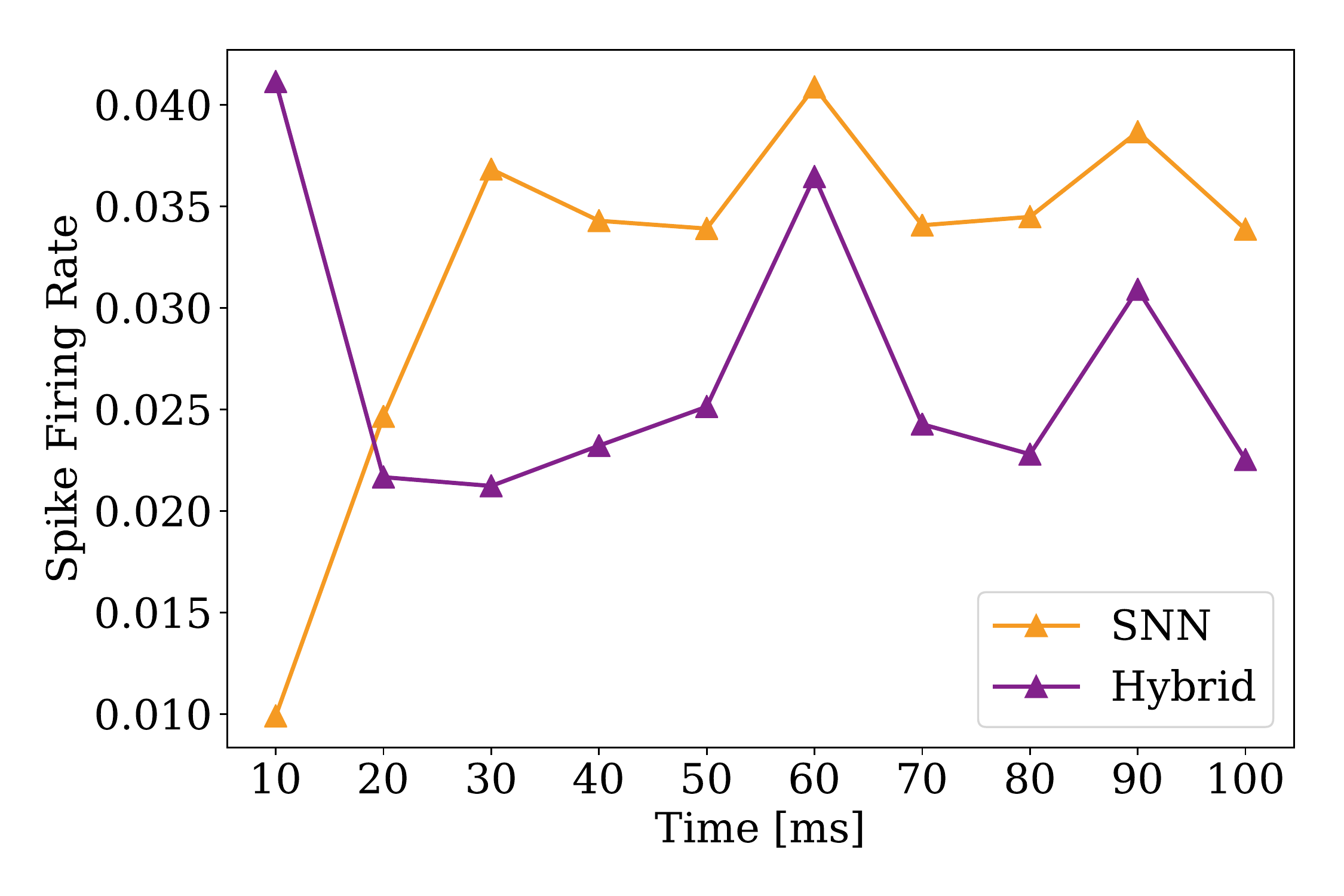}
    \setlength{\belowcaptionskip}{-10pt}
    \caption{\textbf{Effect of state initialization on spike firing rates across time steps.} The SNN consumes 46~mW of energy before state initialization, while energy is decreased to 30~mW after state initialization.}
    \label{fig:spike_firing_rate}
\end{figure}

This section compares spike activity before and after state initialization of spiking neurons.
Fig.~\ref{fig:spike_firing_rate} shows the average spike firing rate throughout the network for each time step. 
The spike rate gradually increases for the SNN experiment as membrane potentials build up and more spikes are emitted over time.
In contrast, spike firing rates in the hybrid model show high firing rates in the first time step due to the initialized states. 
Overall, there is reduced firing activity following state initialization, leading to a 35\% decrease in power consumption from 46~mW to 30~mW.

\subsection{Comparison with State-of-the-Art}
\label{sec:exp:sota}

\subsubsection{2D Pose Estimation}
\label{subsec:results_2d_hpe}

Here we compare the performance of our hybrid approach and its ANN counterpart with previous work on 2D pose estimation.
ANN experiments reported at a specific rate are achieved by inputting events in a sliding window manner.
Tab.~\ref{tab:dhp19_mjpe2d} reports 2D results on the test set for DHP19~\cite{dhp19} and compares MPJPE in pixels on the two camera views used in prior work. 
We compare against Calabrese et al.~\cite{Calabrese19cvprw}, which uses an Hourglass style network to process dense event representations, and Baldwin et al.~\cite{tore_volume}, which reuses the network from~\cite{Calabrese19cvprw} but instead uses specialized TORE volumes as inputs. Both methods use artificial neural networks. 

\begin{figure}
    \centering
    \includegraphics[width=1\columnwidth]{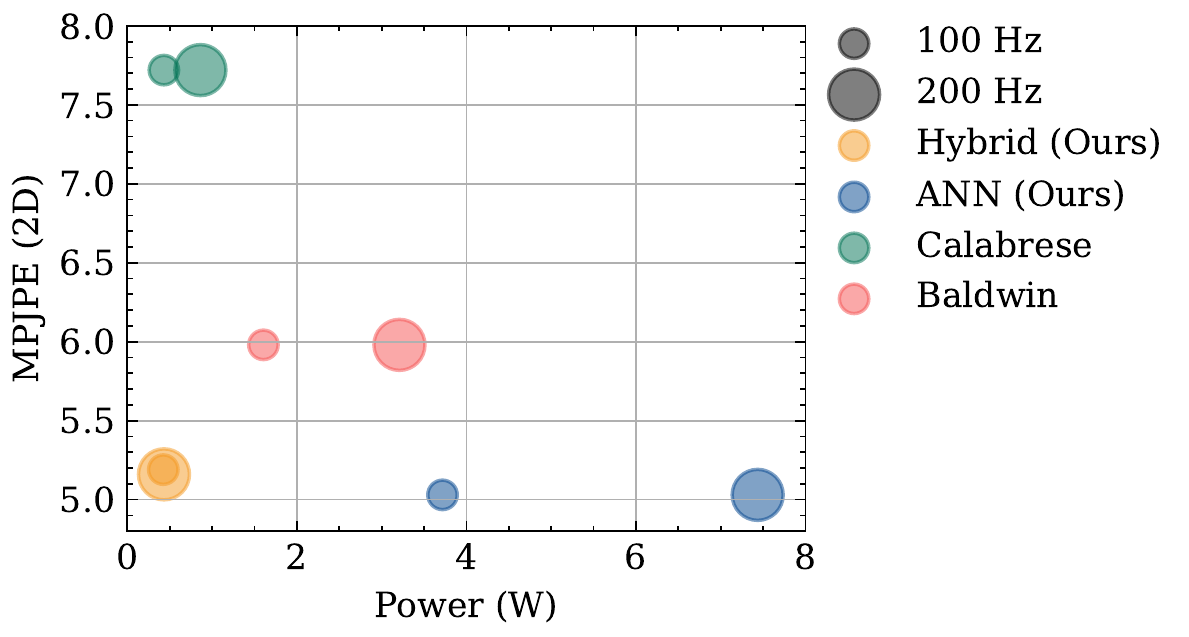}
    \caption{\textbf{2D pose estimation performance vs. power consumption} at 100 and 200 Hz.}
    \label{fig:pow_acc_plot}
\end{figure}

\begin{figure}
    \centering
    \includegraphics[width=1\columnwidth]{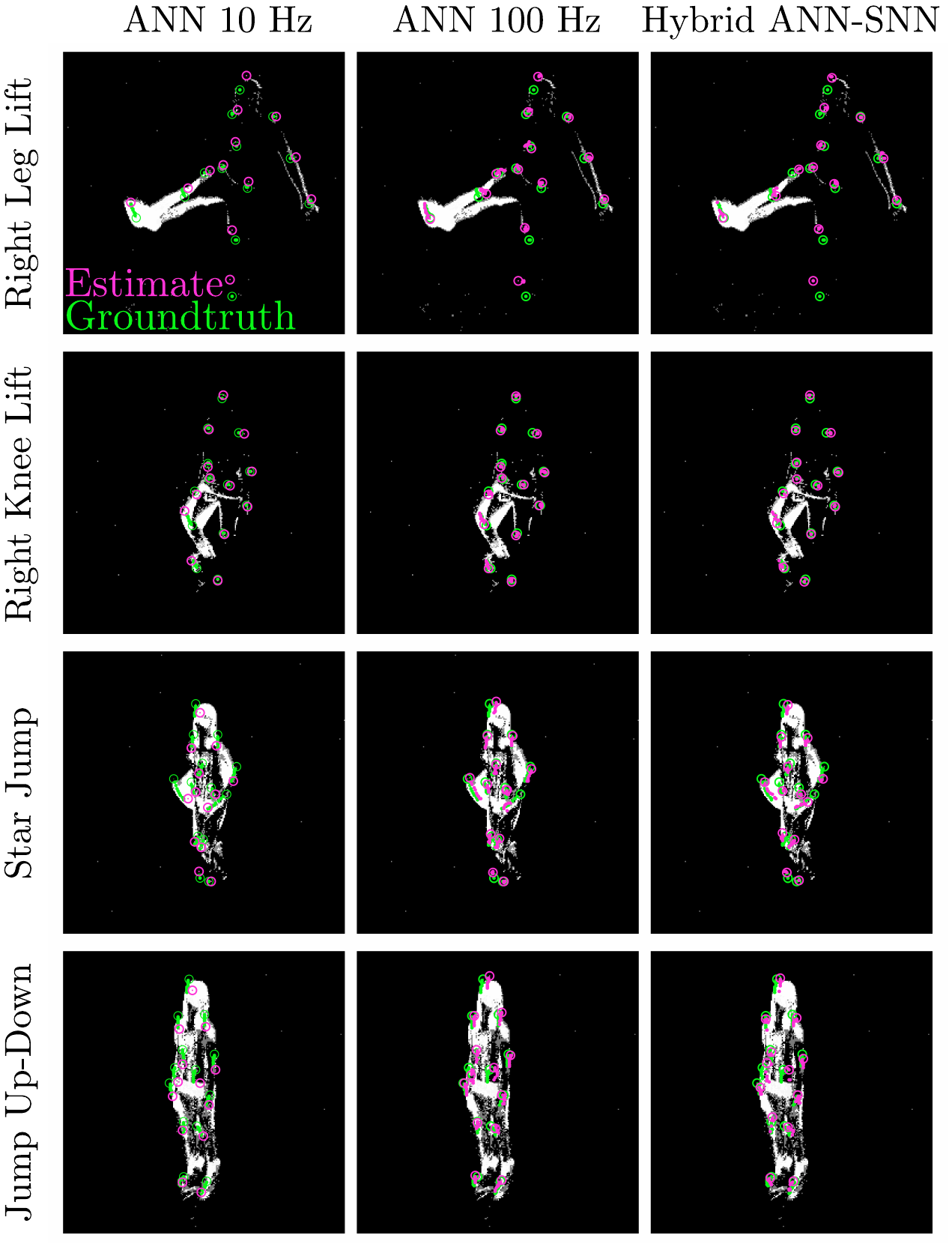}
    \setlength{\belowcaptionskip}{-0.5cm}
    \caption{\textbf{Qualitative 2D pose estimation} for different test subjects and actions. From left to right, samples from ANN at 10 Hz, 100 Hz, and our hybrid ANN - SNN model at 100 Hz with \textcolor{my_green}{green} markers indicating \textcolor{green}{groundtruth} and \textcolor{my_pink}{pink} markers indicating \textcolor{my_pink}{predictions}. Large circles indicate the end prediction of the sequence.}
    \label{fig:2d_qualitative_plot}
\end{figure}

Tab.~\ref{tab:dhp19_mjpe2d} shows that our pure ANN achieves the best score with 5.03 px compared to all previous methods but consumes the most energy with 3.718 W.
In contrast, the hybrid model with 0.424 W is 8x more energy efficient compared to the pure ANN model, with only a 6\% and 3\% decrease in accuracy for camera views 3 and 2, respectively. 
In comparison with previous work, the hybrid model is the most energy efficient while outperforming previous works. 
A key advantage of using SNNs is that outputs are continuous time, meaning that we may increase the discretization step of the SNN to smaller values, i.e., higher rate outputs, without significantly impacting power consumption. Fig.~\ref{fig:pow_acc_plot} visualizes all methods' accuracy and power consumption at different rates. 
The SNN part of our hybrid model only consumes 5\% of the energy. 
Therefore, increasing the update rate allows higher rate predictions with little to no increase in power consumption while maintaining the same accuracy. 
For approaches relying only on ANNs, power consumption increases linearly with the prediction rate. 
From Fig.~\ref{fig:pow_acc_plot}, we see that our hybrid model at 200 Hz retains good accuracy with minimal change in power consumption.
Fig.~\ref{fig:2d_qualitative_plot} visualizes the tracking performance of our approach at 100 Hz for different movements and test subjects, in comparison to ANN at 10 and 100 Hz. 

\begin{table}[t!]
\centering
\caption{\textbf{2D pose estimation performance on DHP19} and energy consumptions. The hybrid model runs the ANN and SNN at 10~Hz and 100~Hz, respectively, and the pure ANN models are run at 100~Hz.
Best values are highlighted in bold, and second best are underlined.}
\label{tab:dhp19_mjpe2d}
\resizebox{\columnwidth}{!}{%
\begin{tabular}{ccccccr}
\toprule
\multirow{2}{*}{\textbf{Method}} & \multirow{2}{*}{\textbf{Model}} & \multicolumn{2}{c}{\textbf{2D MPJPE $\downarrow$}} & \multicolumn{2}{c}{\textbf{\# Ops/s (G)}} & \multirow{2}{*}{\textbf{\begin{tabular}[c]{@{}c@{}}Power \\ (W)\end{tabular}}} \\ \cmidrule(lr){3-4} \cmidrule(lr){5-6}
               &                  & Cam 2      & Cam 3      &  MAC  & AC &               \\ \midrule
Calabrese~\cite{dhp19}         & ANN              & 7.72          & 7.61    &  255 & 0  & 0.431       \\
Baldwin~\cite{tore_volume} & ANN              & 5.98          & 5.25         & 949 & 0 & 1.605          \\
Ours           & ANN              & \textbf{5.03} & \textbf{4.67} & 2200 & 0 & 3.718            \\
Ours           & SNN              & 21.37 & 19.17 &  0.5 & 121 & \textbf{0.046}            \\ 
Ours           & Hybrid & \underline{5.19}          & \underline{4.97}         & 233 & 79  & \underline{0.424} \\ \bottomrule
\end{tabular}%
}
\end{table}

Next, we evaluate our approach on the Event-Human3.6M~\cite{lifting_monocular_events} dataset. 
We perform two experiments, feeding either RGB images or event representation to the ANN. 
This shows our approach's generality to work with multimodal (events+frames) data. The ANN is deployed at 5 Hz, while the SNN uses a discretization step of 10 ms, resulting in 100 Hz updates.
Tab.~\ref{tab:h36m_mjpe2d} reports our scores and power consumption.
Our hybrid approach with RGB images fed to the ANN reports on par accuracy of 4.66 pixels with previous work while being 26.7 times more energy efficient with 0.21 W power consumption compared to the 5.61 W of previous work.
Due to the model complexity of previous work, our hybrid experiments with only events, the last row of Tab.~\ref{tab:h36m_mjpe2d}, fall short of state-of-the-art but show competitive performance.

\begin{table}[t!]
\centering
\vspace*{-\baselineskip}
\caption{\textbf{2D pose estimation scores on Event-Human3.6M} with energy consumptions. The hybrid model runs the ANN and SNN at 5~Hz and 100~Hz, respectively whereas the pure ANN models are run at 100~Hz.}
\label{tab:h36m_mjpe2d}
\resizebox{\columnwidth}{!}{%
\begin{tabular}{ccccccr}
\toprule 
\multirow{2}{*}{\textbf{Method}} & \multirow{2}{*}{\textbf{Model}} & \multirow{2}{*}{\textbf{Modality}} & \multirow{2}{*}{\textbf{2D MPJPE $\downarrow$}} & \multicolumn{2}{c}{\textbf{\# Ops/s (G)}} & \multirow{2}{*}{\textbf{\begin{tabular}[c]{@{}c@{}} Power \\ (W)\end{tabular}}} \\ \cmidrule(lr){5-6}  
&                  &       &     & MAC  & AC &               \\ \midrule
Scarpellini~\cite{lifting_monocular_events} & ANN       & Events       & \underline{4.66} & 3321 & 0 & 5.61             \\
Ours               & ANN       & RGB          & \textbf{4.19} &  2025 & 0 & 3.42              \\
Ours               & ANN       & Events       & 5.09 & 2200 & 0 & 3.72              \\
Ours               & Hybrid & RGB + Events & \underline{4.66} & 108 & 75 & \textbf{0.21}                     \\
Ours               & Hybrid & Events       & 5.76    & 117 & 70 & \underline{0.22}                     \\ \bottomrule

\end{tabular}%
}
\end{table}

\begin{table}[ht]
\centering
\vspace*{-\baselineskip}
\caption{\textbf{3D pose estimation scores on DHP19}. MPJPE is reported in millimeters.}
\label{tab:dhp19_mpjpe3d}
\resizebox{\columnwidth}{!}{%
\begin{tabular}{ccccccr}
\toprule
\multirow{2}{*}{\textbf{Method}} & \multirow{2}{*}{\textbf{Model}} & \multirow{2}{*}{\textbf{Triang.}} & \multirow{2}{*}{\textbf{3D MPJPE $\downarrow$}}   & \multicolumn{2}{c}{\textbf{\# Ops/s (G)}} & \multirow{2}{*}{\textbf{\begin{tabular}[c]{@{}c@{}}Power \\  (W)\end{tabular}}}\\ \cmidrule(lr){5-6} 
               &                  &       &     &  MAC  & AC &               \\ \midrule
Calabrese~\cite{dhp19}         & ANN       & Geom.    & 87.6     & 255 & 0  & \underline{0.431}    \\
Baldwin~\cite{tore_volume} & ANN       & NN & 58.4  & 984 & 0 & 1.664          \\
Ours           & Hybrid & Geom.    & \underline{57.7} & 233 & 79  & \textbf{0.424}     \\ 
Ours           & Hybrid & NN & \textbf{54.2}  & 268 & 79 & 0.483 \\ \bottomrule
\end{tabular}%
}
\end{table}

\vspace{-5pt}
\subsubsection{3D Pose Estimation}\label{subsec:results_3d_hpe}
For 3D HPE on the DHP19 dataset, we use the 2D detection provided by our method and then use two triangulation methods to generate 3D points.
First, we use geometrical triangulation from the two camera views to compare against Calabrese et al.~\cite{dhp19}.
Second, we use neural network-based (learned) triangulation for a fair comparison with Baldwin et al.~\cite{tore_volume}.
Tab.~\ref{tab:dhp19_mpjpe3d} summarizes the results and shows that our approach yields the best trade-off between performance and energy consumption.
Our hybrid approach with geometrical triangulation requires less power than Calabrese et al.~\cite{dhp19} while substantially reducing the MPJPE by 34\%, from 87.6 mm to 57.7 mm.
When using learned triangulation, our method achieves an MPJPE reduction of 4.2~mm compared to Baldwin et al.~\cite{tore_volume}, while at the same time consuming 3.4 times less energy.

\subsection{Limitation Analysis}
\label{sec:exp:limitation}

To enhance the performance of our approach and ensure temporal consistency of predictions, one could consider conditioning the ANN on the prior SNN states. This is a limitation of the current approach but could be overcome with ANN to SNN cross-attention in future work.

%% file: content/ablation_figure.tex
\begin{figure*}[ht]
\centering
  \includegraphics[width=1\textwidth]{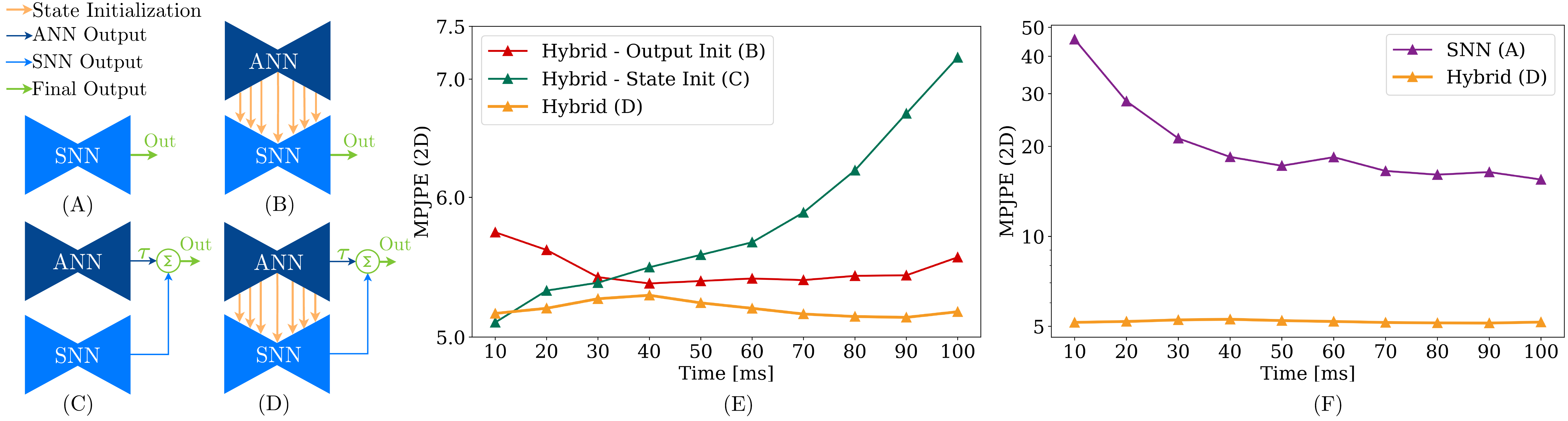}
  \caption[.]{\textbf{Overview of the ablation experiments.} Schematic of (A) pure SNN without state initialization and output initialization, (B) hybrid model without output initialization, (C) hybrid model without state initialization, and (D) our proposed hybrid model with state and output initialization. Plots of accuracy over time of our approach against (E) only output initialization and only state initialization ablated and (F) both ablated. All plots show 2D MPJPE scores on the entire test set for camera view \#2, with initializations performed at $t=0$.}
  \label{fig:method_ablations}
\end{figure*}

%% file: content/05_conclusion.tex
\section{Conclusion}\label{sec:conclusion} %

We presented a hybrid ANN-SNN architecture that delivers fast and precise inference with reduced computational costs.
Our work identified that the slow convergence of SNNs is due to the initialization of spiking neuron membrane states.
Our solution to this problem involves initializing both SNN states and the output state with an ANN.
This approach enables SNNs to immediately produce accurate predictions without requiring a warm-up phase, thereby reducing latency.
Our experimental results demonstrate that this hybrid architecture can reduce energy consumption up to 88\%, with only a 4\% decrease in performance on human pose estimation with respect to a standard ANN. When compared to SNNs, our method achieves a 74\% lower error.
We anticipate that this research will inspire further investigations at the intersection of neuromorphic engineering and conventional deep learning.

%% file: content/appendix.tex
\section{Appendix}

\subsection{Overview}
Here, we provide additional information supporting the main manuscript. In what follows we will refer to Figures, Tables, Sections, and Equations from the main manuscript with the prefix ``M-", and use no prefix for new references in the appendix. We start by providing further analysis of our state initialization scheme (Sec.~\ref{sec:app:init}), then provide additional network details in Sec.~\ref{sec:app:net_details}. We also attach a supplementary video to visualize our low-latency human pose estimation network's output.

\subsection{State Initialization Analysis}
\label{sec:app:init}

\subsubsection{Initialized State Values}
\begin{figure}[bht]
    \centering
    \includegraphics[width=0.93\columnwidth]{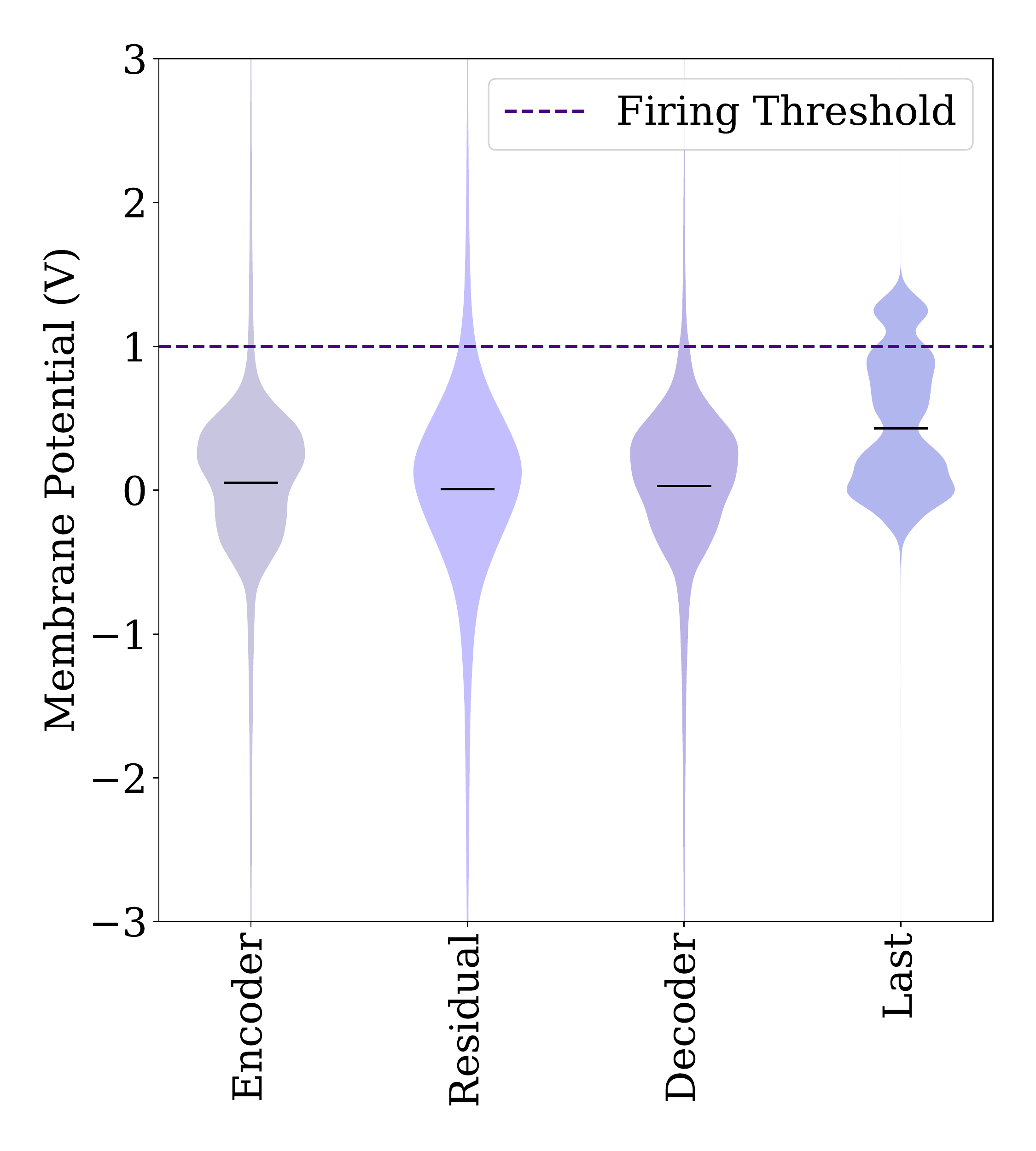}
    \caption{\textbf{Violin plots of membrane potential values after state initialization across layers of the SNN.} Black lines indicate the mean state value at every layer. `Last' indicates the last state initialized layer before the output.}
    \label{fig:spike_firing_rate_appendix}
\end{figure}

In Fig.~\ref{fig:spike_firing_rate_appendix} we show the distribution of initial membrane potentials predicted by our ANN, grouped by the encoder, residual blocks, decoder, and last initialized layer before the output. Note that the firing threshold is 1, meaning that certain neurons are initialized in a firing state. In particular, the output layer shows a high proportion of these kinds of neurons. We call these states that are initialized close to firing, or even in a firing state meta-stable. This meta-stable state is important to reduce latency since it means that few input events can immediately elicit a network response since the membrane potentials are close to firing.
We also see a long tail of inhibited neurons that are initialized with a negative membrane potential.

\subsubsection{Effect of Last Layer State Initialization}\label{sec:app:ellsi}

\begin{figure}[bht]
    \centering
    \includegraphics[width=0.93\columnwidth]{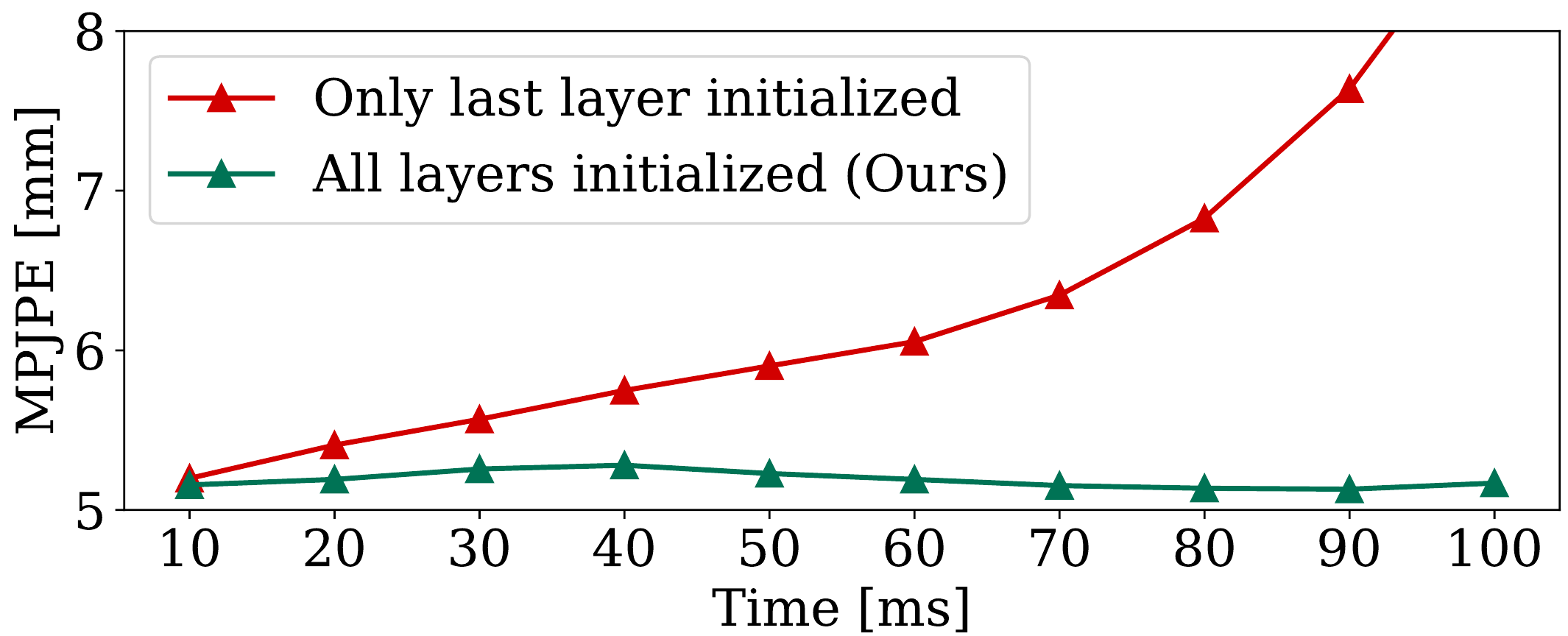}
    \caption{\textbf{Effect of last layer state initialization on performance across time steps.}}
    \label{fig:last_layer_mp}
\end{figure}

We test the impact of the membrane potential initialization in the first layers on the SNN
results by initializing only the last layer with learned membrane potentials and the rest with zeros. 
We see that, in fact, the initial layers influence the SNN performance significantly, as can be seen in Fig.~\ref{fig:last_layer_mp}. 

\subsubsection{Visualization of Last Layer State Initialization}\label{sec:app:vllsi}

\begin{figure}[bht]
    \centering
    \includegraphics[width=0.93\columnwidth]{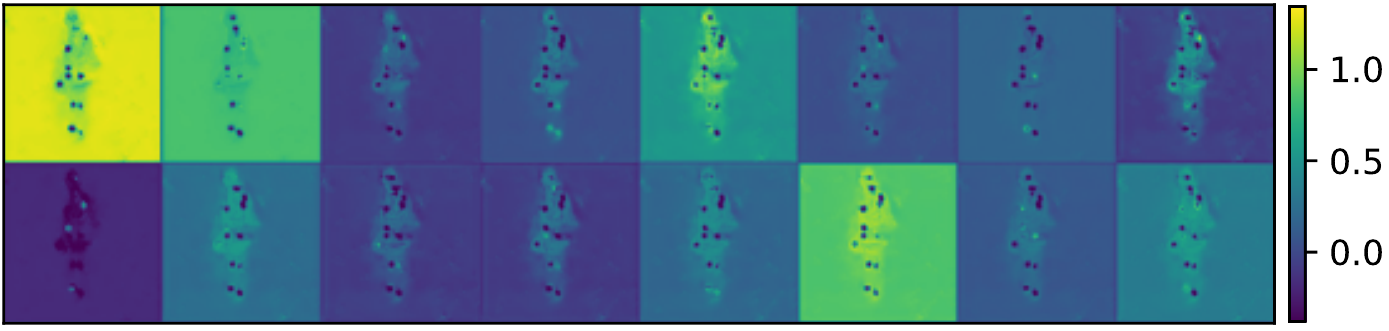}
    \caption{\textbf{Visualization of initialized membrane potentials for multiple last layer channels.}}
    \label{fig:mp_vis}
\end{figure}

In Fig.~\ref{fig:mp_vis}, we plot the initialized membrane potentials of multiple channels of the last layer for additional insights. We observe that each channel raises the membrane potential more in the subject’s projection (foreground) than in the background, leading to a greater likelihood of spikes in the foreground. Moreover, while each channel dampens potential keypoint locations, they tend to activate more intensely at certain keypoints, indicating that channels seem to specialize in specific subsets of keypoints.

\subsection{Network Details}
\label{sec:app:net_details}

The CNN and SNN architecture details are given in Tables~\ref{tab:cnn_arch} and~\ref{tab:snn_arch}, respectively.
For each layer, padding is calculated to preserve spatial dimensions.
Both tables are given with respect to the resolution of the DHP19 dataset, 256x256. 
For the Event-Human3.6M dataset, the resolution is 320x256.

Each convolutional layer in the CNN is followed by a leaky ReLU layer with a negative slope of 0.1.
Columns 1-6, are the encoder layers where an average pooling layer is followed by two convolutions. 
Columns 7-11 are decoder layers, and operations are as follows: (i) interpolation, (ii) convolutional layer, (iii) concatenation with skip connections of the same resolution, and (iv) convolution. 
Finally, the last column is a simple prediction layer with no activation function. 

Each convolutional layer in the SNN is followed by a batch norm, and leaky integrate \& fire neuron layer.
The first column is the spike encoder, columns 2-4 are encoder, 5-6 are residual, and 7-9 are decoder layers. 
Decoder blocks perform concatenation with skip connections at the same spatial resolution and are upsampled together. 
Finally, the last layer is a single convolutional layer.

\FloatBarrier
\begin{table}[h]
\centering
\caption{\textbf{CNN architecture details.} Changes in spatial resolution are due to 2x2 average pooling or bilinear interpolation by a scale of 2. The input channel is of size 20 for event representations or 3 for RGB images.}
\label{tab:cnn_arch}
\resizebox{\linewidth}{!}{%
\begin{tabular}{l|cccccccccccc}
\toprule
\textbf{Layer} &
  \textbf{1} &
  \textbf{2} &
  \textbf{3} &
  \textbf{4} &
  \textbf{5} &
  \textbf{6} &
  \textbf{7} &
  \textbf{8} &
  \textbf{9} &
  \textbf{10} &
  \textbf{11} &
  \textbf{12} \\ \midrule
\textbf{Kernel size}    & 7   & 5   & 3   & 3   & 3   & 3   & 3   & 3   & 3   & 3   & 3   & 3   \\
\textbf{Output channel} & 32  & 64  & 128 & 256 & 512 & 512 & 512 & 256 & 128 & 64  & 32  & 13  \\
\textbf{Output H, W}    & 256 & 128 & 64  & 32  & 16  & 8   & 16  & 32  & 64  & 128 & 256 & 256 \\ \bottomrule
\end{tabular}}%
\end{table}
\FloatBarrier

\begin{table}[H]
\centering
\caption{\textbf{SNN architecture details.} Changes in spatial resolution are due to convolutions with stride 2 and bilinear interpolation of scale 2. The input channel is of size 2.}
\label{tab:snn_arch}
\resizebox{\linewidth}{!}{%
\begin{tabular}{l|cccccccccc}
\toprule
\textbf{Layer} & \textbf{1} & \textbf{2} & \textbf{3} & \textbf{4} & \textbf{5} & \textbf{6} & \textbf{7} & \textbf{8} & \textbf{9} & \textbf{10} \\ \midrule
\textbf{Kernel size}    & 5   & 5   & 5   & 5   & 3   & 3   & 5   & 5   & 5   & 1 \\
\textbf{Stride}         & 1   & 2   & 2   & 2   & 1   & 1   & 1   & 1   & 1   & 1                      \\
\textbf{Output channel} & 32  & 64  & 128 & 256 & 256 & 256 & 128 & 64  & 32  & 13                     \\
\textbf{Output H, W}    & 256 & 128 & 64  & 32  & 32  & 32  & 64  & 128 & 256 & 256                    \\ \bottomrule
\end{tabular}%
}
\end{table}